\pdfoutput=1

\documentclass[11pt]{article}

\usepackage[final]{acl}

\usepackage{times}
\usepackage{latexsym}

\usepackage[T1]{fontenc}

\usepackage[utf8]{inputenc}

\usepackage{microtype}

\usepackage{inconsolata}
\usepackage{pifont}
\usepackage{graphicx}
\usepackage{booktabs}
\usepackage{multirow}
\usepackage{array}
\usepackage{amsmath}
\usepackage{amsfonts}
\usepackage{xcolor}
\usepackage{tcolorbox}
\usepackage{multicol}
\definecolor{querycolor}{HTML}{D9EAF7}  
\definecolor{responsecolor}{HTML}{EAF7D9} 
\definecolor{evalcolor}{HTML}{F7EAD9}  
\definecolor{DarkGreen}{rgb}{0.0, 0.5, 0.0}

\title{Unlocking Recursive Thinking of LLMs: Alignment via Refinement}

\author{
 \textbf{Haoke Zhang\textsuperscript{\ding{168}, \ding{171}}},
 \textbf{Xiaobo Liang\textsuperscript{\ding{168}, \ding{171}}},
 \textbf{Cunxiang Wang\textsuperscript{\ding{169}, \ding{170}}},
 \textbf{Juntao Li\textsuperscript{\ding{168}, \ding{171}} \thanks{\quad Corresponding Author}},
 \textbf{Min Zhang\textsuperscript{\ding{168}, \ding{171}}}
\\
 \textsuperscript{\ding{168}}Soochow University,
 \textsuperscript{\ding{169}}Zhipu AI,
 \textsuperscript{\ding{170}}Tsinghua University \\
 \textsuperscript{\ding{171}} Key Laboratory of Data Intelligence and Advanced Computing, Soochow University \\
\\
 \texttt{hkzhangnlp@stu.suda.edu.cn, \{xbliang, ljt\}@suda.edu.cn}
}

\begin{document}
\maketitle 
\begin{abstract}

The OpenAI o1-series models have demonstrated that leveraging long-form Chain of Thought (CoT) can substantially enhance performance. 
However, the recursive thinking capabilities of Large Language Models (LLMs) remain limited, particularly in the absence of expert-curated data for distillation.
In this paper, we propose \textbf{AvR}: \textbf{Alignment via Refinement}, a novel method aimed at unlocking the potential of LLMs for recursive reasoning through long-form CoT. 
AvR introduces a refinement process that integrates criticism and improvement actions, guided by differentiable learning techniques to optimize \textbf{refinement-aware rewards}.
As a result, the synthesized multi-round data can be organized as a long refinement thought, further enabling test-time scaling.
Experimental results show that AvR significantly outperforms conventional preference optimization methods. 
Notably, with only 3k synthetic samples, our method boosts the performance of the LLaMA-3-8B-Instruct model by over 20\% in win rate on AlpacaEval 2.0.
Our code is available at Github~\footnote{\url{https://github.com/Banner-Z/AvR.git}}.

\end{abstract}

\section{Introduction}

Long-form CoT~\citep{wei2022chain} plays a crucial role in test-time scaling~\cite{snell2024scaling, muennighoff2025s1}, as it enables recursive reasoning akin to human cognitive processes when addressing query intent~\citep{simon1971human, schon1979reflective, bereiter2013psychology}.
However, most existing LLMs lack sequential revision capability, making it difficult to iteratively refine response quality through extended reasoning~\citep{chen2024not, wang2025thoughts}.
Traditional alignment methods optimize LLM outputs solely based on final preference rewards, overlooking critical processes such as reflection and refinement of previously generated content.
Although recent approaches, such as process supervision and reinforcement learning (RL) without supervised fine-tuning~\citep{team2025kimi, team2024deepseek}, have shown notable gains in reasoning performance, the challenge of achieving effective test-time scaling remains an open research question.

\begin{figure}[t]
   \centering
   \vspace{0.2cm}
   \includegraphics[width=1\linewidth]{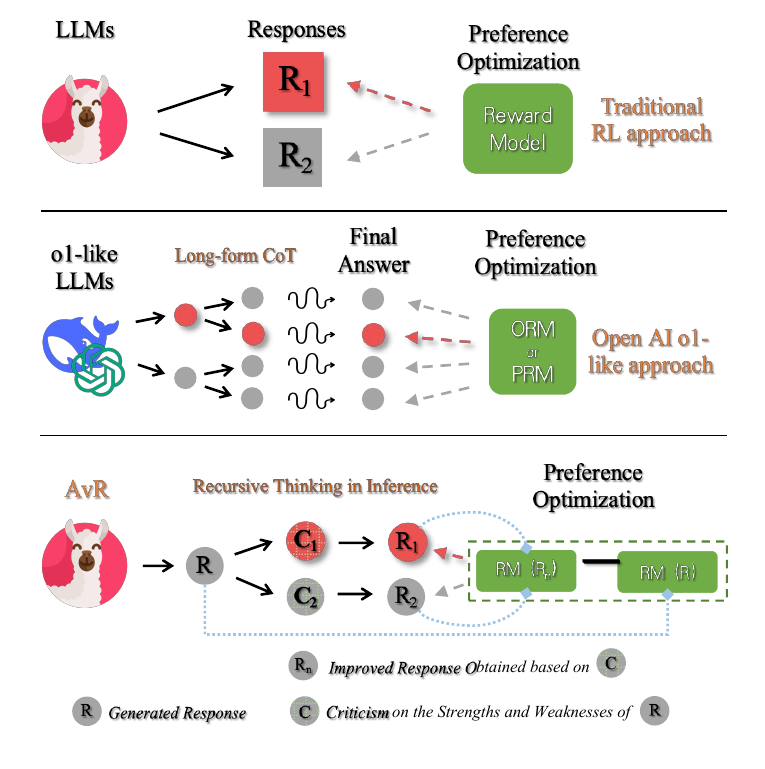}
  \caption{Reward assignment comparison between traditional RL in NLP, o1-like methods, and our AvR.}
  \label{fig:intro}
  \vspace{-0.3cm}
\end{figure}

We begin by comparing how \textbf{reward assignment} is handled across various Reinforcement Learning from Human Feedback (RLHF) algorithms, as illustrated in Figure~\ref{fig:intro}.
Traditional preference optimization methods fine-tune LLM behavior to directly maximize a reward function based on the Bradley–Terry preference model~\citep{schulman2017proximal, rafailov2024direct}.
However, despite its efficiency, this approach fails to capture distinctions among generated responses, often leading to similar errors being repeated during parallel sampling.
In contrast, o1-like models incorporate long-form CoT reasoning to align models with outcome-level or process-level rewards~\citep{lightman2023let, wang2024math}.
As a bonus, by leveraging RL algorithms and test-time computation, LLMs naturally learn to engage in recursive thinking behavior, incorporating self-verification, reflection.
However, activating such capabilities incurs significant computational overhead~\citep{team2024deepseek}, including the requirement for powerful backbone models, extensive sampling, and training costs.
\textbf{\textit{A central challenge is whether their complementary strengths can be integrated to enhance response quality with lower computational cost.}}

In this work, we introduce a refinement-aware reward to unlock recursive thiking capability, which is both effective and efficient for alignment.
Specifically, our method aims to maximize the preference reward of refinement, incorporating \textit{criticism} and \textit{improvement} actions~\cite{kim2023language}.
This approach draws inspiration from differential learning~\citep{sutton1988learning}, which ensures that the decision-making process can be effectively improved by optimizing the reward between different refinements.

In implementation, we design a two-stage framework to synthesize long-form recursive thinking data and optimize LLMs to align with refinement-driven behavior.
\textbf{\textit{In Stage 1}}, we leverage LLMs to perform criticism and improvement actions to refine initial responses via \textit{parallel sampling}, then rank the refined outputs to generate paired preference data.
Then, we apply DPO~\citep{rafailov2024direct} to train LLMs to maximize the refinement-aware reward, enabling the model to learn a policy that favors progressively improved outputs.
\textbf{\textit{In Stage 2}}, we use the Stage 1 model to synthesize high-quality CoT trajectories via \textit{sequential revision}, which serve as guidance to promote long-term refinement behavior, effectively encouraging models to optimize for long-horizon rewards.
We demonstrate that our proposed AvR model, trained on only 10k examples, significantly outperforms the baseline (LLaMA-3-8B-Instruct) and even surpasses RL-based methods trained on 60k samples, as evaluated on AlpacaEval 2 and Arena-Hard v0.1.
Our contributions are as follows:
\begin{itemize}
    \item We introduce a novel \textbf{refinement-aware reward} that enables LLMs to optimize their policy and unlock recursive thinking capabilities, achieving both effectiveness and efficiency.  
    \item Experimental results demonstrate that AvR outperforms current DPO algorithms on Alpaca Eval 2.0, achieving a \textbf{51.0\% Win Rate} and a \textbf{51.4\% LC Win Rate}.  
    \item AvR presents a fundamentally different approach from online RL methods for \textbf{synthesizing high-quality long-form CoT data}, offering fresh insights into test-time scaling.

\end{itemize}

\section{Related Work}

Test-time compute, which leverages inference time resources to refine model outputs, has shown promise in enhancing reasoning quality.
To better understand how models can acquire and benefit from recursive thinking capabilities, we provide a brief review of recent approaches.

\paragraph{Self-Correction}
While this process allows models to refine their outputs through feedback and can be further enhanced by CoT synthesis, recent work has highlighted a key limitation: 
LLMs often struggle to correct their own errors in the absence of an external reward function~\citep{kamoi2024can, zhang2024understanding, jiang2024self}.
However, obtaining high-quality CoT trajectories and reliable reward functions remains challenging in open-domain tasks.
To mitigate the reliance on external reward functions in multi-turn self-improvement, \citet{qu2024recursive} proposed a majority voting mechanism, enabling the model to select improved responses through self-comparison.
Other approaches such as self-rewarding\citep{yuan2024self} and meta-rewarding\citep{wu2024meta} utilize LLMs-as-a-judge to generate internal reward signals, facilitating iterative self-improvement.
However, these methods primarily focus on enhancing LLMs’ ability to evaluate or generate responses in multiple rounds, rather than enabling models to naturally engage in response refinement.

\paragraph{RL-based Methods}
Recently, o1-like LLMs have demonstrated remarkable performance on complex reasoning tasks, particularly in mathematics and code generation~\citep{OpenAIO1, teamqwq, team2024deepseek}.
Monte Carlo Tree Search (MCTS) has proven to be an effective strategy for synthesizing high-quality CoT data across math, code, and general generation tasks~\citep{zhang2024o1, guan2025rstar, zhao2024marco}.
\citet{qin2024o1} proposed a journey learning paradigm that encourages models to go beyond shortcut solutions, promoting complete exploration including trial and error, reflection, and backtracking.
Similarly, \citet{min2024imitate} introduced an imitate, explore, and self-improve framework to reproduce slow-thinking reasoning systems.
\citet{wang2024drt} presented DRT-o1, which applies a long-form CoT process to machine translation and demonstrates strong performance in literature translation.
\citet{kumar2024training} proposed SCoRe, an online reinforcement learning method that significantly enhances LLMs’ self-correction abilities on math and code benchmarks, further showcasing the potential of recursive reasoning.
Despite the effectiveness of these approaches, they often incur substantial computational costs.
In contrast, our method provides an efficient alternative, achieving significant performance improvements with only a small amount of refinement data, thereby offering a more accessible path to unlocking recursive thinking capabilities in LLMs.

\section{Problem Formulation}

Unlike the traditional definition of token-level Markov Decision Processes (MDPs) in language generation tasks, we define both the query and each response from LLMs as independent actions.
We introduce a new action, \textit{refinement}, which can be viewed as a combination of self-judgment and self-correction performed by the LLMs.
Formally, we define the multi-step MDP as a tuple $\{ \mathcal{S}, \mathcal{A}, \mathcal{T}, \mathcal{R}, \gamma \}$, 
where $\mathcal{S}$ represents the state space, with the initial state $s_0$ sampled from the initial prompt distribution $\rho_0$.
The action space $\mathcal{A}$ consists of the possible sequences sampled by LLMs policy $\pi$, including both direct responses and \textit{refinement}. 
The transition function $\mathcal{T}$ is typically deterministic for LLMs, modeled as $\mathcal{P} (s_{t+1}) = [s_t : a_t]$, where the next state is obtained by concatenating the previous state and the selected action. 
The reward function $\mathcal{R}$ provides immediate reward, while the discount factor $\gamma$ balances short-term returns and long-term returns.

Most importantly, traditional token-level MDPs are optimized to maximize rewards by directly generating the final response, with explicit reward signals available only at the final state.
In contrast, our proposed MDP aims to maximize the cumulative reward over multiple steps:
\begin{equation}
   \max_{\pi} \mathbb{E}_{s_{t+1} \sim [s_t : a_t], a_t \sim \pi(\cdot | s_t)} \left [ \sum^{|T|}_{t=0} \gamma^t \mathcal{R} (s_{t+1}, s_t) \right ], \nonumber
\end{equation}
\textbf{where explicit reward signals $\mathcal{R} (s_{t+1}, s_t)$ are provided at every time step, rather than being delayed until the final state.}

Recursive thinking aims to encourage models to receive positive reward signals (i.e., $\mathcal{R} > 0$) at each step of an iterative reasoning process.
For an initial response $s_0$ and a sequence of refinements $\{s_0, s_1, \ldots, s_T\}$, the refinement process can be naturally decomposed into two levels of optimization:
1) Single-step optimization, where the goal is to ensure that a refinement action yields a higher reward than the original response, i.e., $\mathcal{R}(s_1, s_0) > \mathcal{R}(s_0)$.
2) Multi-step optimization, which imposes a stronger constraint: each subsequent refinement step must produce a response with a progressively higher reward than the previous step, i.e.,
$\mathcal{R}(s_{t+1}, s_t) > \mathcal{R}(s_t, s_{t-1})$.
An algorithm that satisfies these conditions would enable LLMs to acquire recursive thinking capabilities, empowering them to iteratively refine their outputs in a self-improving manner.

In particular, we define a \textbf{Refinement-aware Reward} that enforces $s_t$ satisfy:
Each refinement must be better than the initial response $s_0$;
Each refinement must also be an improvement over the previous step.
During optimization, we perform rejection sampling to discard any refinement trajectories that violate this condition:
\begin{align}
\text{Accept}(s_t) = 1 &\quad \text{if } \mathcal{R}(s_{t+1}, s_{t}) > 0 \notag \\
                       &\quad \text{and } \mathcal{R}(s_{t+1}, s_{t}) > \mathcal{R}(s_0) \notag \\
= 0                   &\quad \text{otherwise}. \nonumber
\end{align}
This ensures that the model learns from only effective and cumulatively beneficial refinement steps.

\section{Method}

\begin{figure*}[t]
   \centering
   \includegraphics[width=\textwidth]{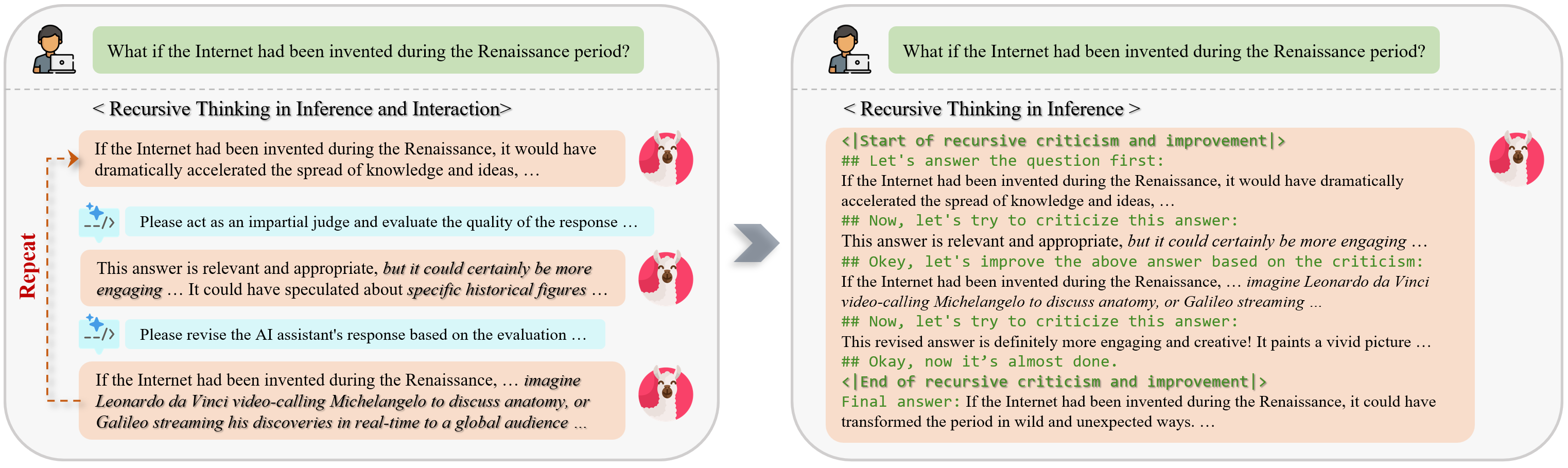}
  \caption{Illustration of two distinct reasoning paradigms.
  The left side depicts a \textbf{multi-step reasoning process}, where each step iteratively refines the previous response through explicit improvement.
  The right side illustrates a \textbf{single-step reasoning process}, wherein the model leverages recursive thinking to complete the user’s query.}
  \label{fig:overview}
  \vspace{-0.3cm}
\end{figure*}

\begin{figure*}[htbp]
  \centering
  \includegraphics[width=\textwidth]{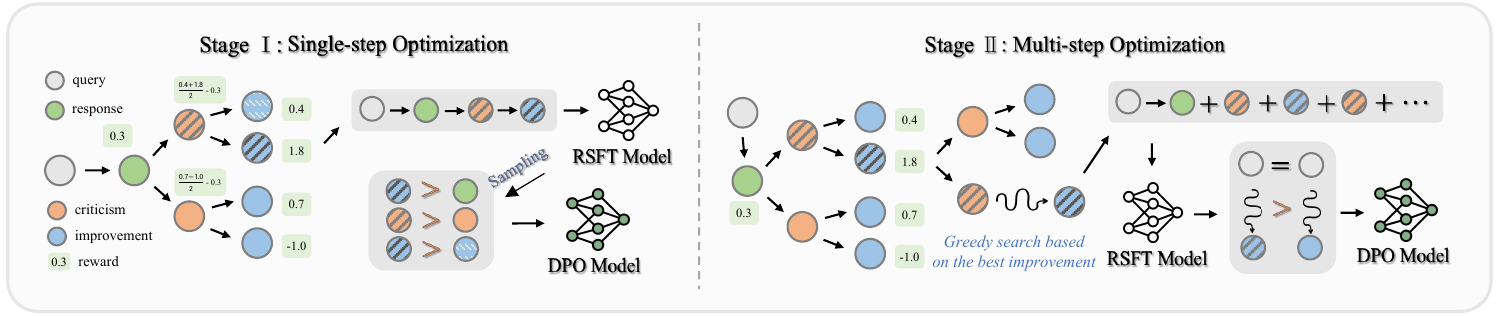}
  \caption{Illustration of our framework: on the left is the first stage, enabling recursive thinking in multi-step inference and interaction; on the right is the second stage, unlocking recursive thinking in inference autonomously.}
  \label{fig:method}
  \vspace{-0.3cm}
\end{figure*}

In this section, we propose a two-stage framework to unlock the recursive thinking capabilities of LLMs.
As shown on the left side of Figure~\ref{fig:overview}, the first stage explicitly guides the model to refine a previously generated response using carefully designed prompts.
In contrast, the right side illustrates the second stage, where the model transitions to autonomous recursive thinking, performing multi-step refinement without relying on external instructions.
The model constructs a recursive thinking trajectory by generating internal control statements such as “Now, let’s try to criticize this answer,” “Okay, let’s improve the above answer based on the criticism,” and “Now it’s almost done.”
These intermediate reasoning steps are enclosed within special tokens to distinguish them from the final response.
Ultimately, the model decides on its own when the refinement process is complete and proceeds to generate the final answer.

\subsection{Stage \uppercase\expandafter{\romannumeral 1}: Single-step optimization}

To enable recursive thinking in multi-step inference for LLMs, we begin by guiding the model to obtain positive reward signals through single-step optimization on each instance.
In particular, we construct a refinement tree composed of \textit{criticism} and \textit{improvement} nodes, which allows the model to explore suitable refinement trajectories.
As illustrated in the left part of Figure~\ref{fig:method}, we score both the original response and each refined response using a Bradley-Terry reward model.
Our goal is to ensure that, along any trajectory within the refinement tree, the model can modify previous content and consistently receive positive reward feedback.

We begin by applying Reject-Sampling Supervised Fine-Tuning (RSFT) on the backone model, using the best improvement trajectory determined by the scores of refined responses.
The RSFT training data consists of multi-turn refinement dialogues, as illustrated in the left part of Figure~\ref{fig:overview}.
Next, we perform joint optimization of the three core reasoning behaviors: \textit{generation}, \textit{criticism}, and \textit{improvement} using Direct Preference Optimization (DPO).
The pairwise training data are constructed from the refinement tree as follows:
For the \textit{generation} step, we pair the best improved response with the original response; samples are discarded if no improvement yields a higher reward;
For the \textit{criticism} step, we select a pair of criticism based on their preference scores;
For the \textit{improvement} step, we choose a pair of candidate improvements under the same criticism node.
An example loss function for the generation step is given below:
\begin{align}
    \label{eqn:dpo}
    \mathcal{L}_{\texttt{DPO}}(\pi_{\theta} ; \pi_{\texttt{ref}}) = & - \mathbb{E}_{ (q, r^+, r^-)  \sim \mathcal{D} } \nonumber\\
    \bigg[ \log \sigma \bigg( \beta \log \frac{\pi_{\theta}(r^+|q)}{\pi_{\texttt{ref}}(r^+|q)} & - \beta \log \frac{\pi_{\theta}(r^-|q)}{\pi_{\texttt{ref}}(r^-|q)} \bigg) \bigg], \nonumber
\end{align}
where $q, r$ is the query and response, $\pi_{\texttt{ref}}$ is typically the RSFT model, $\beta $ is the hyperparameter that controls the proximity of the policy $\pi_{\theta}$  to the original policy $\pi_{\texttt{ref}}$.

\subsection{Stage \uppercase\expandafter{\romannumeral 2}: Multi-step optimization}

To ensure that the model can benefit from test-time compute scaling, we integrate recursive thinking into a long-form CoT reasoning framework, thereby eliminating the reliance on explicit prompts and step-specific supervision.
Building upon Stage \uppercase\expandafter{\romannumeral 1}, we adopt a greedy search strategy to synthesize recursive thinking trajectories, as illustrated on the right side of Figure~\ref{fig:method}.
Specifically, given an initial response or the best refinement from the previous iteration—the model generates x criticisms, each followed by y improvements (step = 2 in our implementation).
Among the resulting candidates, the best improvement is selected using a score computed by the Bradley-Terry reward model, and this serves as the input for the next iteration.
The recursive process continues until no further improvement surpasses the best response from the previous step.
Finally, we train an RSFT model using these automatically selected recursive thinking trajectories, which correspond to the highest-ranked responses under the learned preference model.

In addition, we observe that the reward model tends to favor longer outputs, which often leads the RSFT model to generate unnecessarily verbose responses during inference.
To mitigate this issue and enable the model to produce concise yet high-quality outputs when appropriate, we introduce a length-controlled DPO fine-tuning stage on top of the RSFT model.

Concretely, we perform multi-sample inference using the RSFT model (five samples per input in practice), and identify the highest-scoring and lowest-scoring responses according to the reward model.
We then filter the samples to retain only those where the highest-scoring response is shorter than the lowest-scoring one.
This subset is subsequently used to perform DPO training, thereby encouraging the model to prefer concise outputs when they are also more preferable under the reward model.

\section{Experiments}

\subsection{Experimental Settings}
\paragraph{Implementation Details}
LLaMA-Factory\footnote{\url{https://github.com/hiyouga/LLaMA-Factory.git}} is employed in our training. To fetch a broad range of baselines, our experiments are conducted on Meta-Llama-3-8B-Instruct\footnote{\url{https://huggingface.co/meta-llama/Meta-Llama-3-8B-Instruct}}~\citep{llama3modelcard}. We use Skywork-Reward-Gemma-2-27B-v0.2\footnote{\url{https://huggingface.co/Skywork/Skywork-Reward-Gemma-2-27B-v0.2}}~\citep{liu2024skywork} as the Bradley-Terry Model throughout the work.
We utilize llama3-ultrafeedback armorm\footnote{\url{https://huggingface.co/datasets/princeton-nlp/llama3-ultrafeedback-armorm}} dataset, which uses 60k prompts from Ultrafeedback~\citep{cui2023ultrafeedback} and regenerate the chosen and rejected response pairs with Meta-Llama-3-8B-Instruct.
To enhance the process, we leverage Qwen2.5-32B-Instruct-GPTQ-Int8\footnote{\url{https://huggingface.co/Qwen/Qwen2.5-32B-Instruct-GPTQ-Int8}}~\citep{qwen2.5} as a corrector to produce criticisms and improvements for the training of RSFT model in \textbf{AvR Stage \uppercase\expandafter{\romannumeral 1} model}. All data synthesis following the completion of this RSFT model is performed using our own models, without the introduction of any external models.
We set the temperature to 0.7 and top\_p to 0.8 for all training data generation and testing. The learning rate is 5e-6 for SFT and 5e-7 for DPO. The value of $\beta$ is set to 0.01. The batch size is 64, and the number of training epochs is 1.0. The cutoff length is set to 2048 for \textbf{AvR Stage \uppercase\expandafter{\romannumeral 1} model} and 8192 for \textbf{AvR Stage \uppercase\expandafter{\romannumeral 2} model}. 
We utilize the DeepSpeed~\cite{rasley2020deepspeed} library, Zero Redundancy Optimizer (ZeRO)~\cite{rajbhandari2020zero} Stage 3, and FlashAttention~\cite{dao2023flashattention}, along with a mixed precision computation approach using bfloat16~(BF16) and tfloat32~(TF32), across 8 NVIDIA A100 GPUs. 

\paragraph{Evaluation} We evaluate our models on two open-ended conversation benchmarks: Alpaca Eval 2~\citep{alpaca_eval}, and Arena-Hard v0.1~\citep{li2024crowdsourced}. 
Alpaca Eval 2 is an LLM-based automatic evaluation with 805 questions from 5 datasets. The model's responses are compared with those of GPT-4-Turbo. The win rate represents the likelihood that the auto-evaluator favors the evaluated model's responses. To mitigate the length bias in the auto-evaluator, the Length-controlled Win Rate is utilized. 
Arena-Hard v0.1 contains 500 challenging user queries sourced from Chatbot Arena, comparing the models' responses against a baseline model~(GPT-4-0314). 
Following the standard experimental setup, GPT-4 Turbo~(corresponding to GPT-4-Preview-1106) is used as the auto-evaluator in three benchmarks.

\begin{table*}[t]
    \centering
    \resizebox{1\textwidth}{!}
    {
    \begin{tabular}{@{}l|c|cccc@{}}
        \toprule
        \textbf{Method} & \textbf{Init Model} & \textbf{Data scale} & \parbox{2.5cm}{\centering Win Rate} & \parbox{2.5cm}{\centering Len.-control. Win Rate} & \parbox{1.5cm}{\centering Length} \\
        \midrule
        SEED & Llama-3-8B-Ins & - & $25.0\%$ & $25.0\%$ & 1956 \\
        \textit{+refine} & Llama-3-8B-Ins & - & $22.4\%_{\color{red}{\downarrow 2.6\%}}$ & $21.4\%_{\color{red}{\downarrow 3.6\%}}$ & 1952 \\
        \textit{+refine} & Qwen2.5-32B-Ins & - & $33.3\%_{{\color{black}{\uparrow 8.3\%}}}$ & $34.7\%_{{\color{black}{\uparrow 9.7\%}}}$ & 1943 \\
        gpt-4-0613 & gpt-4-0613 & - & $15.8\%$ & $30.2\%$ & -\\
        Llama-3.1-405B-Ins & Llama-3.1-405B-Ins & - & $39.1\%$ & $39.3\%$ & - \\
        \midrule
        \textbf{RL Methods} \\
        DPO & Llama-3-8B-Ins &60k& $37.9\%_{{\color{black}{\uparrow 12.9\%}}}$ & $40.3\%_{{\color{black}{\uparrow 15.3\%}}}$ & - \\
        KTO & Llama-3-8B-Ins &60k& $31.8\%_{\color{black}{\uparrow 6.8\%}}$ & $33.1\%_{\color{black}{\uparrow 8.1\%}}$ & - \\
        ORPO & Llama-3-8B-Ins &60k& $37.8\%_{\color{black}{\uparrow 12.8\%}}$ & $41.1\%_{\color{black}{\uparrow 16.1\%}}$ & - \\
        SimPO & Llama-3-8B-Ins &60k& $40.5\%_{\color{black}{\uparrow 15.5\%}}$ & $\underline{44.7\%}_{\color{black}{\uparrow 19.7\%}}$ & 1825 \\
        \midrule
        \textbf{Meta-Rewarding LLM} \\
        \textit{Iteration 1} & Llama-3-8B-Ins &5k & $27.6\%_{\color{black}{\uparrow 2.6\%}}$ & $27.9\%_{\color{black}{\uparrow 2.9\%}}$ & 1949 \\ 
        \textit{Iteration 2} & Llama-3-8B-Ins &5k~(10k)& $33.3\%_{\color{black}{\uparrow 8.3\%}}$ & $32.7\%_{\color{black}{\uparrow 7.7\%}}$ & 2001 \\ 
        \textit{Iteration 3} & Llama-3-8B-Ins &5k~(15k)& $37.2\%_{\color{black}{\uparrow 12.2\%}}$ & $35.5\%_{\color{black}{\uparrow 10.5\%}}$ & 2064 \\ 
        \textit{Iteration 4} & Llama-3-8B-Ins &5k~(20k)& $39.5\%_{\color{black}{\uparrow 14.5\%}}$ & $39.4\%_{\color{black}{\uparrow 14.4\%}}$ & 2003 \\ 
        \midrule
        \textbf{AvR Stage \uppercase\expandafter{\romannumeral 1}} \\
        RSFT & Llama-3-8B-Ins & 10k & $21.1\%_{\color{red}{\downarrow 3.9\%}}$ & $22.1\%_{\color{red}{\downarrow 2.9\%}}$ & 1925 \\
        \textit{+refine round 1} & Llama-3-8B-Ins & - & $25.7\%_{\color{black}{\uparrow 0.7\%}}$ & $22.8\%_{\color{red}{\downarrow 2.2\%}}$ & 2234 \\
        \textit{+refine round 2} & Llama-3-8B-Ins & - & $27.2\%_{\color{black}{\uparrow 2.2\%}}$ & $22.6\%_{\color{red}{\downarrow 2.4\%}}$ & 2457 \\
        DPO & Llama-3-8B-Ins & 10k~(20k) & $37.0\%_{\color{black}{\uparrow 12.2\%}}$ & $36.2\%_{\color{black}{\uparrow 11.2\%}}$ & 2179  \\
        \textit{+refine round 1} & Llama-3-8B-Ins & - & $49.2\%_{\color{black}{\uparrow 24.2\%}}$ & $39.1\%_{\color{black}{\uparrow 14.1\%}}$ & 2562  \\
        \textit{+refine round 2} & Llama-3-8B-Ins & - & $\underline{50.8\%}_{\color{black}{\uparrow 25.8\%}}$ & $35.5\%_{\color{black}{\uparrow 10.5\%}}$ & 2963 \\
        \midrule
        \textbf{AvR Stage \uppercase\expandafter{\romannumeral 2}} \\
        RSFT & Llama-3-8B-Ins & 10k & $\textbf{51.0\%}_{\color{black}{\uparrow 26.0\%}}$ & $42.5\%_{\color{black}{\uparrow 17.5\%}}$ & 2687 \\
        \textit{+length control} & Llama-3-8B-Ins & 4k~(14k) & $49.0\%_{\color{black}{\uparrow 24.0\%}}$ & $\textbf{51.4\%}_{\color{black}{\uparrow 26.4\%}}$ & 1989 \\
        \bottomrule
    \end{tabular}
}
    \caption{\label{tab:main-results}
        The experimental results on Alpaca Eval 2, with \textbf{bold numbers} indicating the best performance and \underline{underlined numbers} representing the second-best performance.
      }
\end{table*}

\paragraph{Baselines} 
In our experiments, we primarily introduce three types of baselines. The first consists of well-pretrained and aligned dialogue models, such as Qwen2.5-32B-Instruct-GPTQ-Int8, Llama-3.1-405B-Instruct\footnote{\url{https://huggingface.co/meta-llama/Llama-3.1-405B-Instruct}}, and GPT-4-0613. The second includes models trained with hybrid reinforcement learning methods, including DPO~\citep{rafailov2023direct}, KTO~\citep{ethayarajh2024kto}, ORPO~\citep{hong2024reference}, R-DPO~\citep{park2024disentangling}, and SimPO~\citep{meng2025simpo}. The third category encompasses self-improvement approaches, including Self-Rewarding LLM~\citep{yuan2024self} and Meta-Rewarding LLM~\citep{wu2024meta}.

\subsection{Main Results}
Table~\ref{tab:main-results} presents the main results of our experiments on Alpaca Eval 2. We report both the win rate and the length-controlled win rate to assess the performance of the methods. For each score, we indicate the difference between the results before and after applying the method. Additionally, we provide the average string length of the model outputs and the scale of training prompts for each model\footnote{To mitigate the substantial decline in generation capabilities caused by training with a fixed-format prompting dataset during the AvR Stage \uppercase\expandafter{\romannumeral 1}, we incorporate all 60k responses generated by Meta-Llama-3-8B-Instruct into the RSFT training corpus, thereby preserving its generative abilities.}~(The numbers in brackets represent the cumulative amount of training prompts used if the model was trained iteratively.). We analyze the performance from the following perspectives:

\paragraph{AvR Stage \uppercase\expandafter{\romannumeral 2} model shows significant improvements by constructing high-quality training data through inference scaling.} While the Llama-3-8B-Instruct model struggles with self-correction on Alpaca Eval 2, the win rate on the first generation improves by 12\% after the RSFT and DPO training on only 20k data and the self-correction ability improves even further, resulting in an additional 12.2\% improvement in the quality of the first responses.

\paragraph{LLMs are highly effective at recursive thinking in inference.} With a well-trained AvR Stage \uppercase\expandafter{\romannumeral 1} model, we find that just 10k prompts can lead to a 26\% improvement in the win rate and a 17\% improvement in the length-controlled win rate by constructing long-form CoT data, which encourages recursive criticism and improvement during the inference process. In comparison to meta-rewarding approaches or hybrid reinforcement learning methods, which search for a better response to guide LLMs in generating improved outputs, our method—focused on teaching LLMs the iterative process—yields a greater overall improvement in final output with fewer training data and lower training costs.

\paragraph{Constructing preference pairs by maximizing the difference before and after refinement significantly enhances the self-iterative ability of LLMs.} When comparing the results of RSFT with those of DPO in our AvR Stage \uppercase\expandafter{\romannumeral 1} model, the improvement after one round of refinement is only 4.6\% for RSFT, whereas Qwen2.5-32B-Instruct achieves an 8.3\% improvement, as it generates the training data for RSFT. The DPO model reverses this trend, even though it only uses the data generated by the RSFT model. We believe the key advantage of our DPO approach lies in allowing the LLMs to learn how to maximize the difference between the outputs before and after refinement.

\begin{table}[t]
    \centering
    \resizebox{\linewidth}{!}{
        \begin{tabular}{l|cc|c}
        \toprule
        \textbf{Method} & \textbf{Score} & \textbf{95\% CI} & \textbf{Length} \\
        \midrule
        gpt-3.5-turbo-0125 & 23.3\% & (-2.2, 1.9) & - \\
        gpt-4-0613 & 37.9\% & (-2.8, 2.4) & - \\
        \textbf{Llama-3-8B-Instruct~(Seed)} & 20.6\% & (-2.0, 1.8) & 2485 \\
        \midrule
        \textbf{RL Method} \\
        DPO & 32.6\% & - & - \\
        ORPO & 25.8\% & - & - \\
        R-DPO & 33.1\% & - & - \\
        SimPO & 33.8\% & - & - \\
        \midrule
        \textbf{Self-Rewarding LLM} \\
        \textit{Iteration 1} & 23.2\% & (-1.7, 1.9) & 2438 \\
        \textit{Iteration 2} & 26.3\% & (-2.1, 2.3) & 2427 \\
        \textit{Iteration 3} & 28.2\% & (-2.0, 1.9) & 2413 \\
        \textit{Iteration 4} & 27.3\% & (-2.0, 2.2) & 2448 \\
        \midrule
        \textbf{Meta-Rewarding LLM} \\
        \textit{Iteration 1} & 25.1\% & (-1.9, 1.8) & 2395 \\
        \textit{Iteration 2} & 27.4\% & (-2.0, 2.0) & 2416 \\
        \textit{Iteration 3} & 27.6\% & (-2.3, 2.6) & 2501 \\
        \textit{Iteration 4} & 29.1\% & (-2.3, 2.1) & 2422 \\
        \midrule
        \textbf{AvR Stage \uppercase\expandafter{\romannumeral 2}} & \textbf{34.5\%} & (-2.5, 2.3) & 3144 \\
        \bottomrule
        \end{tabular}
    }
    \caption{The experimental results on Arena Hard v0.1.}
    \vspace{-0.3cm}
    \label{tab:arena-hard}
\end{table}

\paragraph{The AvR Stage \uppercase\expandafter{\romannumeral 2} model requires only a slight DPO to learn length control, relying on self-generated preference data.} Our AvR Stage \uppercase\expandafter{\romannumeral 2} model tends to increase the output length in each refinement round, which results in a lower score on the length-controlled win rate. However, our experiments demonstrate that the AvR Stage \uppercase\expandafter{\romannumeral 2} model can efficiently learn length control by training with just 4k preference pairs. This method achieves an 8.9\% improvement in the length-controlled win rate and reduces the length of the final outputs, bringing them closer to the seed model, with only a 2.0\% decline in the win rate.

\subsection{Results on Arena Hard v0.1}

To further validate the effectiveness of our proposed method, we conduct experiments on the Arena Hard v0.1 benchmark using our AvR Stage \uppercase\expandafter{\romannumeral 2} model. Among the methods trained based on Llama-3-8B-Instruct, AvR Stage \uppercase\expandafter{\romannumeral 2} model achieves the best performance, surpassing that of GPT-3.5-turbo-0125. The 95\% confidence interval~(95\% CI) for our model's performance ranges from~(-2.5, 2.3). In comparison to approaches such as Self-Rewarding LLM and Meta-Rewarding LLM, which require multiple iterations of training, and methods like DPO and SimPO, which rely on extensive preference data for reinforcement learning, our framework demonstrates superior efficiency. Notably, our approach achieves optimal performance with only 10k data points during SFT highlighting its effectiveness and simplicity.

\begin{figure}[t]
   \centering
   \includegraphics[width=\linewidth]{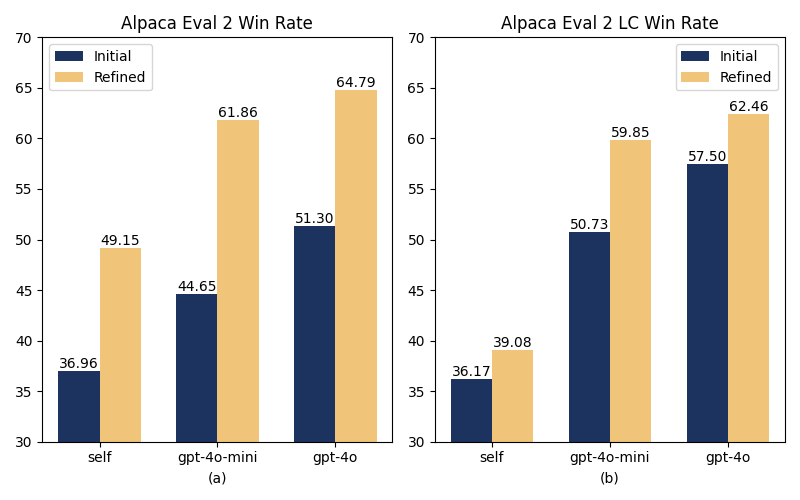}
  \caption{The results demonstrate our model's ability to correct the outputs of more powerful LLMs.}
  \label{fig:correct_stronger_llms}
  \vspace{-0.4cm}
\end{figure}

\subsection{Analysis of Refinement Capability}\label{section:correction}
To further investigate the refinement capability of our AvR Stage \uppercase\expandafter{\romannumeral 1} model, we employ it to refine the outputs of stronger LLMs and evaluate their performance on Alpaca Eval 2. Specifically, we use the original outputs of GPT-4o and GPT-4o-mini\footnote{The GPT-4o-0513 and GPT-4o-mini-2024-07-18 models are utilized in this work.}, as provided by the benchmark, and apply our model to refine these outputs. 
As shown in Figure~\ref{fig:correct_stronger_llms}, our model significantly improves the performance of both models, even though the win rate and length-controlled win rate of our self-refinement results are notably lower than the original outputs of GPT-4o. When combined with the results in Table~\ref{tab:main-results}, it is evident that surpassing the generation quality of powerful models like GPT-4o and GPT-4o-mini is challenging through reinforcement learning or iterative training focused solely on model generation capabilities. However, by enhancing the refinement ability of the model through inference scaling and difference maximization training, we can obtain improved responses from these powerful models. This also suggests promising directions for the future development of model capabilities.

\begin{figure}[t]
   \centering
   \includegraphics[width=\linewidth]{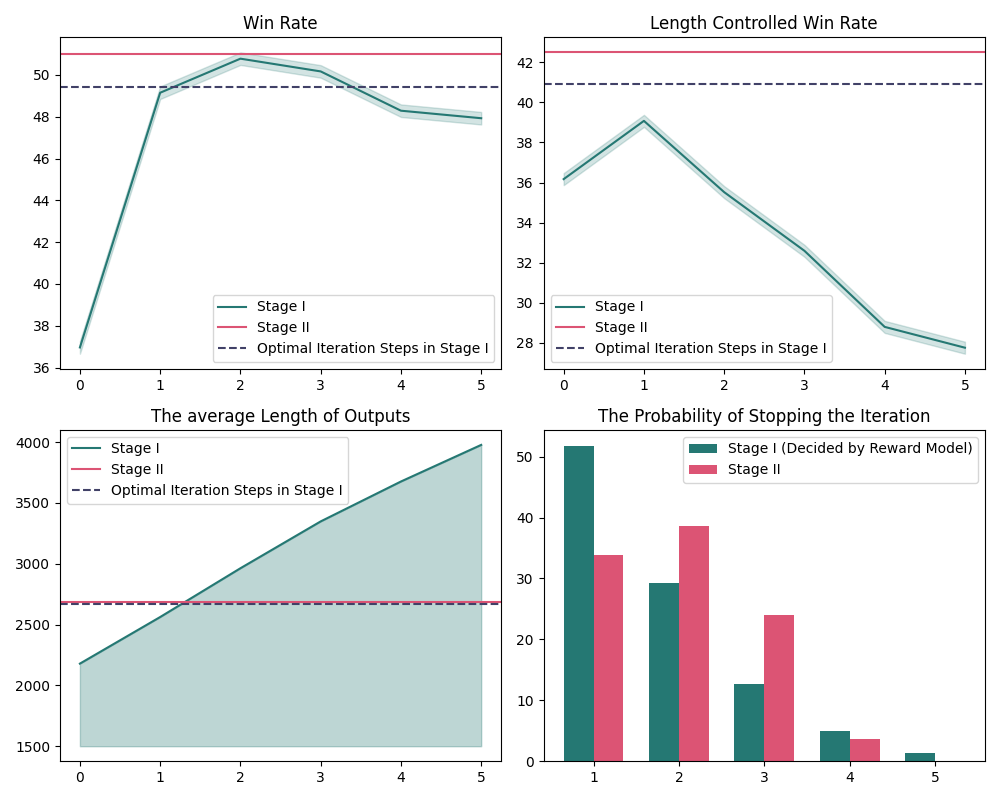}
  \caption{Analysis of iteration capability by comparing the models of the two stages.}
  \label{fig:iteration}
  \vspace{-0.3cm}
\end{figure}

\subsection{Analysis of Iteration Capability}

In Figure~\ref{fig:iteration}, we illustrate the performance variations across multiple iterations, comparing the AvR Stage \uppercase\expandafter{\romannumeral 1} model with the AvR Stage \uppercase\expandafter{\romannumeral 2} model. In our experiments, the AvR Stage \uppercase\expandafter{\romannumeral 1} model is used to generate criticisms and improved responses iteratively, while the Bradley-Terry Model~(Skywork-Reward-Gemma-2-27B-v0.2) acts as a verifier, determining when to stop the iteration. Specifically, the AvR Stage \uppercase\expandafter{\romannumeral 1} model halts iteration when the Bradley-Terry Model detects no improvement in performance compared to the current iteration, and the previous response is then selected as the final output. The four smaller subfigures in the main figure represent the following from top to bottom and left to right: the change in win rate score, the change in length-controlled win rate score, the change in length, and the distribution of the best iteration round.
One key observation is that the AvR Stage \uppercase\expandafter{\romannumeral 1} model fails to prevent the continuous increase in length during the iteration process, although the overall quality does not improve significantly after the first two iterations. This trend is also reflected in the length-controlled win rate and the distribution of optimal iteration rounds. The dotted line in the figure indicates the result of selecting the best iteration round at the instance level. It can be seen that the AvR Stage \uppercase\expandafter{\romannumeral 2} model~(represented by the red solid line) consistently outperforms this configuration on AvR Stage \uppercase\expandafter{\romannumeral 1} model, further demonstrating the effectiveness and importance of unlocking the recursive thinking of LLMs. Additionally, we observe changes in the distribution of iteration rounds. The best iteration round typically falls between the first and third rounds, with the AvR Stage \uppercase\expandafter{\romannumeral 2} model showing a stronger preference for the second round.

\begin{figure}[t]
   \centering
   \includegraphics[width=\linewidth]{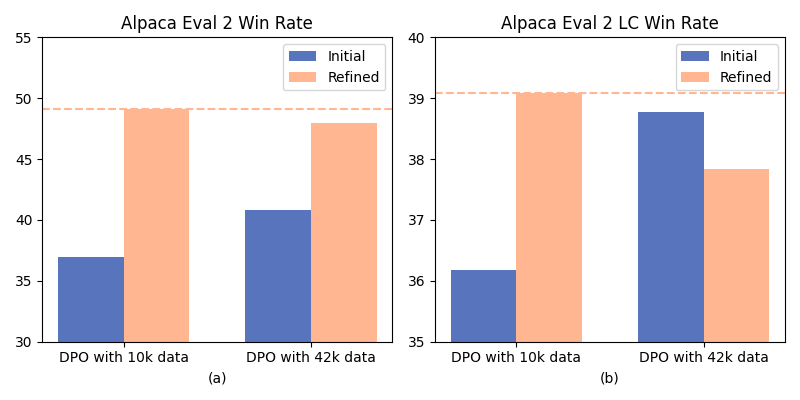}
  \caption{Experimental results on the scaling of DPO training data in the first stage.}
  \label{fig:dpo_data_scaling_win_rate_comparison}
\end{figure}

\begin{table}[t]
    \centering
    \resizebox{1\linewidth}{!}{
        \begin{tabular}{l|cc|c}
        \toprule
        \textbf{Data scale} & \textbf{Win Rate} & \textbf{LC Win Rate} & \textbf{Length}\\
        \midrule
        Init Model & 24.99\% & 24.96\% & 1956 \\
        \midrule
        3k & 45.22\% & 38.69\% & 2683 \\
        6k & 47.71\% & 39.06\% & 2719 \\
        10k & \underline{51.02\%} & \underline{42.46\%} & 2687 \\ 
        20k & \textbf{51.07\%} & \textbf{42.69\%} & 2559 \\ 
        \bottomrule
        \end{tabular}
    }
    \caption{Experimental results on the scaling of SFT training data in the second stage.}
    \vspace{-0.3cm}
    \label{tab:sft-data-scaling}
\end{table}

\subsection{Scaling on Training Data}
In our experiment, we also analyze the scale of the training data, focusing primarily on the DPO training during the Stage \uppercase\expandafter{\romannumeral 1} and the SFT training during the Stage \uppercase\expandafter{\romannumeral 2}. Figure~\ref{fig:dpo_data_scaling_win_rate_comparison} presents the results of the scaling experiment during the DPO training. It is evident that the model’s generation ability significantly improved when trained with 42k prompts compared to the 10k prompts setting. This observation is also reflected in Table~\ref{tab:main-results}—training with 60k prompts resulted in better generation performance than our DPO model. However, the model trained with 42k prompts demonstrates significantly lower correction performance than the model trained with 10k prompts, even leading to a decrease in the length-controlled win rate during the refinement phase. Overall, after one round of refinement, the 10k model outperforms the 42k model. Since our experiment places greater emphasis on the desirable property of self-refinement~(as discussed in Section~\ref{section:correction}), we use the model trained with 10k prompts for all subsequent experiments.
Table~\ref{tab:sft-data-scaling} presents the experimental results on data scaling during the SFT training in the Stage \uppercase\expandafter{\romannumeral 2}. We found that training with just 3k long CoT data resulted in a performance improvement of over 20\%, which was comparable to the performance of the Meta-Rewarding LLM trained on 20k data and the reinforcement learning method trained on 60k data. Up to 10k training data, data scaling consistently led to significant improvements in model performance. However, the improvement from 10k to 20k training data is marginal. We attribute this phenomenon to a combination of model capacity limitations, data quality constraints~(partly due to the misalignment of the reward model), and the training algorithm. In all other experiments, we used the 10k training version.

\section{Conclusion}

In this paper, we propose a novel framework to unlock the recursive thinking capabilities of LLMs.
By leveraging inference-time scaling and maximizing a refinement-aware reward, our method synthesizes high-quality data that supports long CoT reasoning and enables self-improvement within the model itself.
Crucially, our two-stage training and inference framework significantly reduces the GPU memory footprint and training cost compared to RL-based methods.
Experimental results demonstrate that our approach substantially enhances both the generative and corrective abilities of the base model. 
Remarkably, with only a small amount of training data, our method empowers an 8B model to dynamically refine its outputs during inference. 
Furthermore, we observe that our model can effectively refine responses generated by more powerful models, offering new perspectives for improving LLM performance through recursive self-improvement.

\section*{Limitations}
Since the evaluation of dialogue tasks often has problems such as inconsistency and uncertainty, the ability and preference of the reward model often affect the quality of synthetic data and thus the effect of the model. Better reward modeling, such as using LLM-as-a-judge for fine-grained evaluation to provide better criticism for refinement~\citep{wang2023pandalm, liang2024fennec, ankner2024critiqueoutloudrewardmodels}, or other better forms of feedback, will bring more expansion and improvement to our work. In addition, this work does not explore more reinforcement learning methods. We have preliminarily discovered in the experiment that reinforcement learning has significantly enhanced this inference method. We believe this is a direction worthy of further expansion in the future.

\section*{Acknowledgments}

We want to thank all the anonymous reviewers for their valuable comments. This work was supported by the National Science Foundation of China (NSFC No. 62206194), the Natural Science Foundation of Jiangsu Province, China (Grant No. BK20220488), the Young Elite Scientists Sponsorship Program by CAST (2023QNRC001). We also acknowledge MetaStone Tech. Co. for providing us with the software, optimisation on high performance computing and computational resources required by this work.

\bibliography{custom}

\appendix

\section{Appendix}
\label{sec:appendix}

\subsection{Comparison with LongCoT Model}
To compare the o1-like LongCoT model with our method, we evaluate the DeepSeek-R1-Distill-Llama-8B\footnote{\url{https://huggingface.co/deepseek-ai/DeepSeek-R1-Distill-Llama-8B}} model on the Alpaca Eval 2.0 benchmark. As shown in Table~\ref{tab:cot_models}, we observe that DeepSeek-R1-Distill-Llama-8B, trained on large-scale LongCoT data, performs poorly on this benchmark. Moreover, although the distilled DeepSeek R1 model can generate very long thinking, the thinking length of the DeepSeek-R1-Distill-Llama-8B model on Alpaca Eval 2.0 is only 2.8 times that of the final response, while our models can reach 3.5 times or even more than 4 times.

\begin{table*}[t]
    \centering
    \resizebox{1\textwidth}{!}{
    \begin{tabular}{@{}l|cc|cc@{}}
        \toprule
        \textbf{Model} & \parbox{2.5cm}{\centering Win Rate} & \parbox{2.5cm}{\centering Len.-control. Win Rate} & \parbox{3cm}{\centering Length of Final Response} & \parbox{2.5cm}{\centering Length of Thinking}\\
        \midrule
        Meta-Llama-3-8B-Instruct & 25.0\% & 25.0\% & 1956 & - \\
        DeepSeek-R1-Distill-Llama-8B & 22.1\% & 27.9\% & 1615 & 4472 \\
        \midrule
        \textbf{AvR Stage \uppercase\expandafter{\romannumeral 2}} & \\
        RSFT & \textbf{51.0\%} & 42.5\% & 2687 & 11781 \\
        +length control & 49.0\% & \textbf{51.4\%} & 1989 & 6947 \\
        \bottomrule
    \end{tabular}}
    \caption{\label{tab:cot_models}
        The experimental results of our methods and the common approach of using long CoT for reasoning enhancement on Alpaca Eval 2, with \textbf{bold numbers} indicating the best performance.
      }
\end{table*}

\subsection{Experiment on Math Tasks}
We conduct experiments on the Qwen2.5-7B-Instruct\footnote{\url{https://huggingface.co/Qwen/Qwen2.5-7B-Instruct}} model to evaluate performance on several math benchmarks. To refine the model’s responses, we leverage DeepSeek-R1, effectively simulating a well-trained Stage \uppercase\expandafter{\romannumeral 1} model. We sample 1,000 questions from OpenThoughts-114k and employ the Qwen2.5-Math-RM-72B\footnote{\url{https://huggingface.co/Qwen/Qwen2.5-Math-RM-72B}} as our reward model. As shown in Table~\ref{tab:math_results}, our method achieves consistent improvements across various math benchmarks, demonstrating the effectiveness of the proposed approach for math tasks.

\begin{table*}[t]
\centering
\resizebox{1\textwidth}{!}{
\begin{tabular}{@{}l|cccccc@{}}
\toprule
\textbf{Method} & \textbf{Data Scale} & \textbf{AIME24} & \textbf{MATH 500} & \textbf{Olympiad Bench} & \textbf{Minerva Math} & \textbf{GSM8K} \\
\midrule
Qwen2.5-7B-Instruct & - & 10.0 & 74.2 & 36.7 & 33.5 & 91.6 \\
SFT with Our Stage \uppercase\expandafter{\romannumeral 2} Data & 1k & \textbf{16.7} & \textbf{74.8} & \textbf{38.5} & \textbf{34.9} & \textbf{92.4} \\
\bottomrule
\end{tabular}}
\caption{\label{tab:math_results}
Results on math benchmarks for the Qwen2.5-7B-Instruct model with and without our Stage \uppercase\expandafter{\romannumeral 2} data.}
\end{table*}

\subsection{Prompts and Cases}
In this section, we present the prompts and formats used for model training and inference, and provide several cases of the inference process.

\begin{tcolorbox}[colback=querycolor!20, colframe=darkblue, title=Prompts in AvR]
\textbf{User:} \{User Query\} \\
\textbf{Assistant}: \{Response\} \\ 
\textbf{User}: Please act as an impartial judge and evaluate the quality of the response provided by an AI assistant to the user question displayed above. Your evaluation should consider factors such as the helpfulness, relevance, accuracy, depth, creativity, and level of detail of the response. Your evaluation should focus on the assistant's answer to the last user question. Begin your evaluation by providing a short explanation. Be as objective as possible. After providing your explanation, you must rate the response at the end of your answer on a scale of 1 to 10 by strictly following this format: "[[rating]]", for example: "Rating: [[5]]". \\
\textbf{Assistant}: \{Judgement\} \\
\textbf{User}: Please revise the AI assistant's response based on the evaluation provided above, addressing any shortcomings mentioned in the review. Your revision should focus solely on improving the assistant's answer to the last user question. Provide the revised response directly, without any additional commentary. \\
\textbf{Assistant}: \{Refinement\}
\end{tcolorbox}


\begin{table*}[h]
\centering
\scalebox{0.95}{
\begin{tabular}{ p{14cm} }
\toprule
\textbf{User:} Who was the first lead vocal for the rock band Starship? \\
\\
\textbf{Assistant}: The first lead vocalist for the rock band Starship was Marty Balin, who was a founding member of the band. He was the lead singer on many of their early hits, including "Don't Count Me Out," "We Built This City," and "Miracles." \\ 
\\
\textbf{User}: Please act as an impartial judge and evaluate the quality of the response ... \\
\\
\textbf{Assistant}: The AI assistant provided a clear and accurate response to the user's question about the first lead vocalist for Starship. The information given is relevant and directly addresses the user's query. The response is detailed enough to mention specific songs and the founding member's role, which adds depth to the answer without being overly verbose. The information provided is helpful for someone seeking to understand the early lead vocals of the band. The response is also creative in the sense that it includes specific examples of songs and emphasizes the founding member's role, which enhances the user's understanding of the context.
\newline
Rating: [[8]] 
\newline
The rating reflects the high quality of the response, with a minor deduction due to the absence of additional context or historical context that might have further enriched the answer, but this was not necessary given the straightforward nature of the question. The information provided is comprehensive and directly answers the user's query effectively. \\
\\
\textbf{User}: Please revise the AI assistant's response based on the evaluation provided above, ... \\
\\
\textbf{Assistant}: The first lead vocalist for the rock band Starship was Marty Balin, a founding member who played a pivotal role in shaping the band's sound. During the early years, he was the primary lead singer on hits like "Don't Count Me Out" (1969) and "We Built This City" (1979). These songs showcased his distinctive vocals, setting the foundation for the band's later success. Balin's contributions as a lead vocalist are significant in understanding Starship's early years and development. \\
\bottomrule
\end{tabular}
}
\caption{Case on AvR Stage \uppercase\expandafter{\romannumeral 1} Model.}
\label{case1}
\end{table*}

\begin{table*}[h]
\centering
\scalebox{0.95}{
\begin{tabular}{ p{14cm} }
\toprule
\textbf{User:} Please give me a list of planets in our solar system.  I am going to choose which one I want to know more. \\
\textbf{Assistant}: "\textbf{<|Start of recursive criticism and improvement|>}
\newline
\textbf{\#\# Let's answer the question first:}
\newline\newline
Here is the list of planets in our solar system:
1. Mercury
2. Venus
3. Earth
4. Mars
5. Jupiter
6. Saturn
7. Uranus
8. Neptune

Which one would you like to know more about?
\newline\newline
\textbf{\#\# Now, let's try to criticize this answer:}
\newline\newline
**Rating: [[8]]**
The AI assistant provided a concise and relevant list of planets in our solar system, directly addressing the user's request. The response is helpful, accurate, and detailed enough to give the user a clear choice. ...

The only aspect that prevents a perfect score is the absence of additional context or explanation about the solar system structure or the format of the response, which might have added depth without compromising clarity. Given the straightforward nature of the request, this minor omission does not detract significantly from the overall quality. Therefore, the rating reflects a high level of helpfulness and relevance, with a slight deduction for the potential depth that could have been included without compromising clarity. 

**Rating: [[8]]** (The response is highly relevant, accurate, and helpful, with a minor deduction for the absence of additional context without compromising clarity.)
\newline\newline
\textbf{\#\# Okey, let's improve the above answer based on the criticism:}
\newline\newline
Here is the list of planets in our solar system, organized for easy selection:
**1.** **Mercury**: Closest to the Sun, with extreme temperatures and a highly elliptical orbit.
**2.** **Venus**: Hottest planet, shrouded in thick clouds, with surface temperatures reaching 462°C (863°F).

...

Which planet would you like to explore further? (Choose one for a detailed dive!)
\newline\newline
\textbf{\#\# Now, let's try to criticize this answer:}
\newline\newline
**Rating: [[8]]**
The AI assistant provided a comprehensive and structured list of the planets in our solar system, catering to the user's request for a detailed exploration of a chosen planet. The response:
1. **Helpfulness**: The list is detailed and well-organized, offering a clear overview of each planet, making it easy for the user to select one for further exploration.

...

Overall, the response is highly effective in meeting the user's needs and setting the stage for a detailed exploration of their chosen planet.
\newline\newline
\textbf{\#\# Okay, now it’s almost done.}
\newline
\textbf{<|End of recursive criticism and improvement|>}
\newline\newline
Final answer:
Here is the list of planets in our solar system, organized for easy selection:
**1.** **Mercury**: Closest to the Sun, with extreme temperatures and a highly elliptical orbit.
**2.** **Venus**: Hottest planet, shrouded in thick clouds, with surface temperatures reaching 462°C (863°F).
**3.** **Earth**: Our home planet, supporting life, with diverse ecosystems and a moderate climate.

...

Which planet would you like to explore further? (Choose one for a detailed dive!)" \\
\bottomrule
\end{tabular}
}
\caption{Case on AvR Stage \uppercase\expandafter{\romannumeral 2} model.}
\label{tab:case2}
\end{table*}

\end{document}